\documentclass[10pt,twocolumn,letterpaper]{article}

\usepackage{cvpr}              

\usepackage{graphicx}
\usepackage{amsmath}
\usepackage{amssymb}
\usepackage{booktabs}

\usepackage{algorithm}
\usepackage{algorithmic}
\usepackage{wrapfig}
\usepackage{amsthm,amsmath,amssymb}
\usepackage{multirow}
\usepackage{color}
\usepackage{makecell}
\usepackage{mathrsfs}
\usepackage{colortbl}
\usepackage{xcolor}

\definecolor{Red}{RGB}{255,160,122}
\definecolor{Yellow}{RGB}{255,227,132}

\usepackage[pagebackref,breaklinks,colorlinks]{hyperref}

\usepackage[capitalize]{cleveref}
\crefname{section}{Sec.}{Secs.}
\Crefname{section}{Section}{Sections}
\Crefname{table}{Table}{Tables}
\crefname{table}{Tab.}{Tabs.}

\def\cvprPaperID{6625} 
\def\confName{CVPR}
\def\confYear{2023}

\begin{document}

\title{Feature Representation Learning with Adaptive Displacement Generation and Transformer Fusion for Micro-Expression Recognition}

\author{Zhijun Zhai\textsuperscript{1},
{Jianhui Zhao\textsuperscript{1}}\thanks{Corresponding author.}, 
Chengjiang Long\textsuperscript{2}, 
Wenju Xu\textsuperscript{3},
Shuangjiang He\textsuperscript{4},
Huijuan Zhao\textsuperscript{4}
 \\
\textsuperscript{1}{School of Computer Science, Wuhan University, Wuhan, Hubei, China} \\
\textsuperscript{2}{Meta Reality Labs, Burlingame, CA, USA} \\
\textsuperscript{3}{OPPO US Research Center, InnoPeak Technology Inc, Palo Alto, CA, USA} \\
\textsuperscript{4}{FiberHome Telecommunication Technologies Co., Ltd, Wuhan, Hubei, China} \\
{\tt\small zhijunzhai@whu.edu.cn, jianhuizhao@whu.edu.cn, clong1@meta.com, wenjuxu123@gmail.com}
}
\maketitle

\begin{abstract}
   Micro-expressions are spontaneous, rapid and subtle facial movements that can neither be forged nor suppressed. They are very important nonverbal communication clues, but are transient and of low intensity thus difficult to recognize. Recently deep learning based methods have been developed for micro-expression (ME) recognition using feature extraction and fusion techniques, however, targeted feature learning and efficient feature fusion still lack further study according to the ME characteristics. To address these issues, we propose a novel framework Feature Representation Learning with adaptive Displacement Generation and Transformer fusion (FRL-DGT), in which a convolutional Displacement Generation Module (DGM) with self-supervised learning is used to extract dynamic features from onset/apex frames targeted to the subsequent ME recognition task, and a well-designed Transformer Fusion mechanism composed of three Transformer-based fusion modules (local, global fusions based on AU regions and full-face fusion) is applied to extract the multi-level informative features after DGM for the final ME prediction. The extensive experiments with solid leave-one-subject-out (LOSO) evaluation results have demonstrated the superiority of our proposed FRL-DGT to state-of-the-art methods.
\end{abstract}

\section{Introduction}
\label{sec:intro}
\begin{figure}[t]
  \vspace{-0.10cm}
  \centering
   \includegraphics[width=1.0\linewidth]{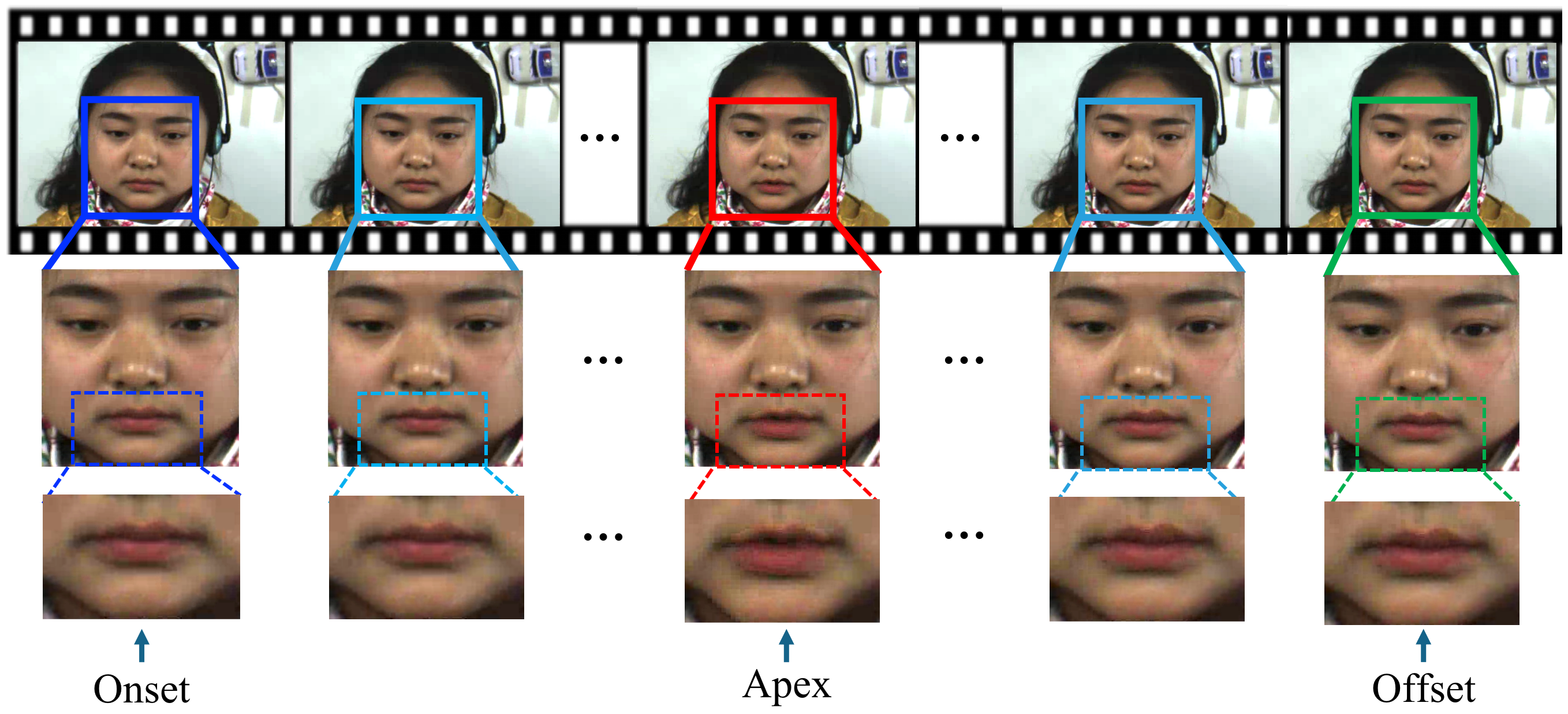}
   \vspace{-0.75cm}
   \caption{A video sequence depicting the order of which onset, apex and offset frames occur. Sample frames are from a ``surprise” sequence in CASME II. Our goal is to design a novel feature representation learning method based on an onset-apex frame pair for facial ME recognition. (Images from CASME II \copyright{Xiaolan Fu})}
   \label{fig:introduction}
   \vspace{-0.45cm}
\end{figure}

As a subtle and short-lasting change, micro-expression (ME) is produced by unconscious contractions of facial muscles and lasts only 1/25th to 1/5th of a second, as illustrated in Figure~\ref{fig:introduction}, revealing a person's true emotions underneath the disguise~\cite{shen2012effects,ekman1969nonverbal}. The demands for ME recognition technology are becoming more and more extensive~\cite{Benjamin2022,lewinski2014predicting}, including multimedia entertainment, film-making, human-computer interaction, affective computing, business negotiation, teaching and learning, {\em etc}. Since MEs have involuntary muscle movements with short duration and low intensity in nature, the research of ME is attractive but difficult~\cite{ben2021video,li2020joint}. Therefore, it is crucial and desired to extract robust feature representations to conduct ME analyses.

A lot of feature representation methods are already available including those relying heavily on hand-crafted features with expert experiences~\cite{liong2018less,fleet2006optical,bilen2016dynamic} and deep learning techniques~\cite{takalkar2018survey,verma2019learnet,van2019capsulenet}.
However, the performance of deep learning networks is still restricted for ME classification, mainly due to the complexity of ME and insufficient training data~\cite{liu2022graph,zhang2021review}. Deep learning methods can automatically extract optimal features and offer an end-to-end classification, but in the existing solutions, dynamic feature extraction is only taken as a data preprocessing strategy. It is not integrated with the subsequent neural network, thus failing to adapt the generated dynamic features to a specific training task, leading to redundancy or missing features. Such shortcoming motivates us to design a dynamic feature extractor to adapt the subsequent ME recognition task.



\begin{figure*}[t]
  \centering
  \vspace{-0.15cm}
   \includegraphics[width=0.95\linewidth]{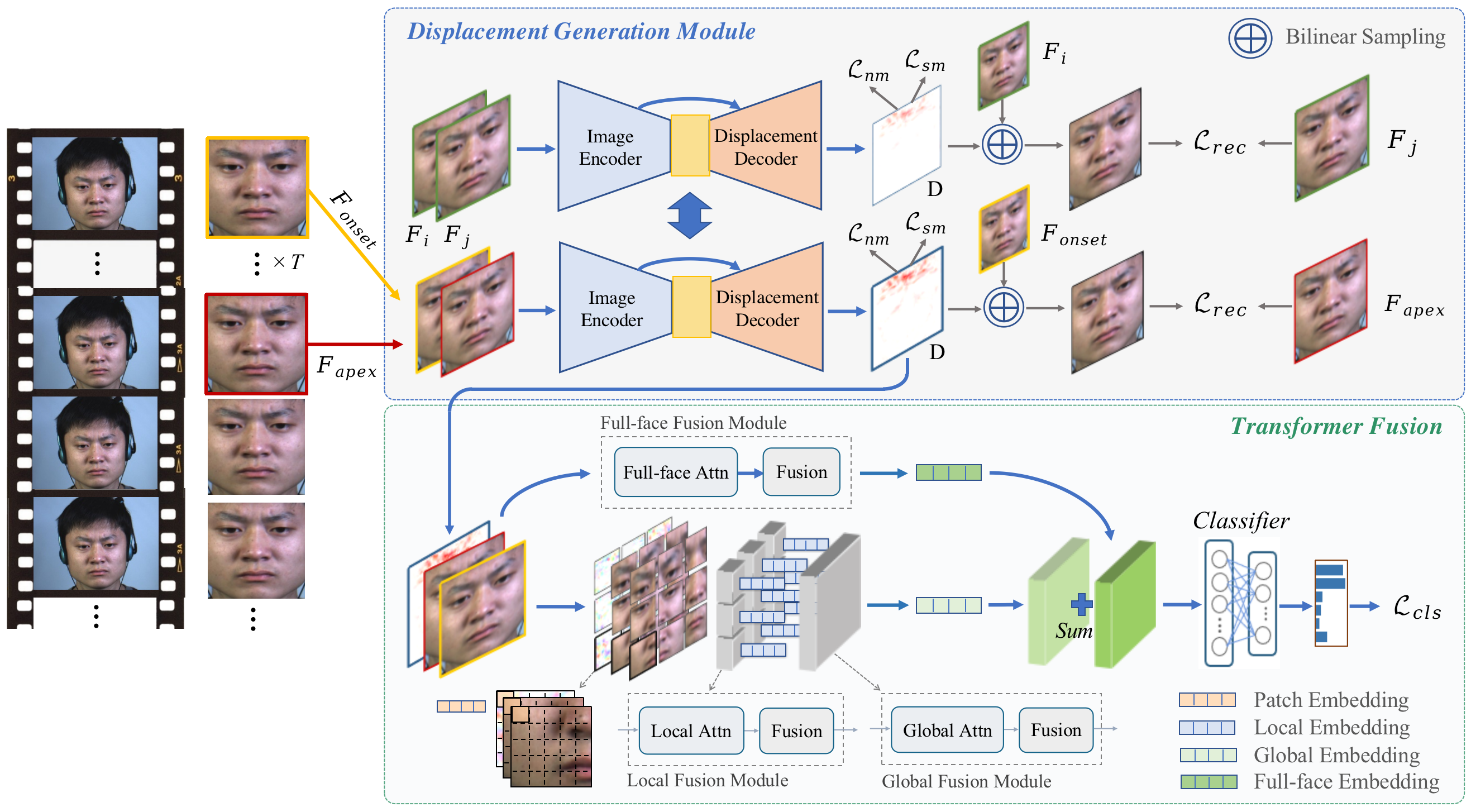}
   \vspace{-0.2cm}
   \caption{The overview of FRL-DGT which is an end-to-end structure. Given an input video clip, we firstly crop the front face regions and then select an onset-apex pair of frames as the input of Displacement Generation Module (DGM) to calculate the displacement adaptively. The Transformer Fusion composed of three modules, {\em i.e.}, local fusion module, global fusion module, and full-face fusion module, \textcolor{black}{takes the onset-apex pair and the displacement between them} as input to extract the final feature representation via both patch and integral feature fusion with Transformer for classification on ME categories. In addition, the random pairs of frames are used as auxiliary data to self-supervise the training of DGM. (Images from CASME II \copyright{Xiaolan Fu})}
   \label{fig:Overall_structure}
   \vspace{-0.35cm}
\end{figure*}

In this paper, we propose a novel end-to-end feature representation learning framework named FRL-DGT, which is constructed with a well-designed adaptive Displacement Generation Module (DGM) and a subsequent Transformer Fusion mechanism composed of the Transformer-based local, global, and full-face fusion modules, as illustrated in 
Figure~\ref{fig:Overall_structure}, for ME feature learning to model global information while focusing on local features. Our FRL-DGT only requires a pair of images, {\em i.e.}, the onset and the apex frames, which are extracted from the frame sequence.



Unlike the previous methods which extract optical flow or dynamic imaging features, our DGM and corresponding loss functions are designed to generate the displacement between expression frames, using a convolution module instead of the traditional techniques. The DGM is involved in training with the subsequent ME classification module, and therefore its parameters can be tuned based on the feedback from classification loss to generate more targeted dynamic features adaptively. 
We shall emphasize that the labeled training data for ME classification is very limited and therefore the supervised data for our DGM is insufficient. To handle this case, we resort to a self-supervised learning strategy and sample sufficient additional random pairs of image sequence as the extra training data for the DGM, so that it is able to fully extract the necessary dynamic features adaptively for the subsequent ME recognition task.

Regarding fusing the dynamic features extracted from DGM, we first adopt the AU (Action Unit) region partitioning method from FACS (Facial Action Coding System)~\cite{Zhang2022} to get 9 AUs, and then crop the frames and their displacements into blocks based on the 9 AUs and the full-face region as input to the Transformer Fusion. We argue that the lower layers in the Transformer Fusion should encode and fuse different AU region features in a more targeted way, while the higher layers can classify MEs based on the information of all AUs. We propose a novel fusion layer with attentions as a linear fusion before attention mechanism~\cite{Yu:TOG2021}, aiming at a more efficient and accurate integration of the embedding vectors. The fusion layers are interleaved with Transformer’s basic blocks to form a new multi-level fusion module for classification to ensure it to better learn global information and long-term dependencies of ME.

To summarize, our main contributions are as follows:
\begin{itemize}
    \item We propose a novel end-to-end network FRL-DGT which fully explores AU regions \textcolor{black}{from onset-apex pair and the displacement between them }to extract comprehensive features via Transformer Fusion mechanism with the Transformer-based local, global, and full-face feature fusions for ME recognition.
    \item Our DGM is well-trained with self-supervised learning, and makes full use of the subsequent classification supervision information in the training phase, so that the trained DGM model is able to generate more targeted ME dynamic features adaptively.
    \item We present a novel fusion layer to exploit the linear fusion before attention mechanism in Transformer for fusing the embedding features at both local and global levels with simplified computation.
    \item We demonstrate the effectiveness of each module with ablation study and the outperformance of the proposed FRL-DGT to the SOTA methods with extensive experiments on three popular ME benchmark datasets.
\end{itemize}

\section{Related Works}

{\bf Micro-expression Recognition} related technologies~\cite{takalkar2018survey,ben2021video,li2020joint,li2022deep,xie2020overview} mainly fall into two types. One category~\cite{liong2018less} (\eg, EMRNet~\cite{liu2019neural}, STSTNet~\cite{liong2019shallow} and DSSN~\cite{khor2019dual}) uses only onset and apex frames. Dynamic features (\eg, optical flow~\cite{fleet2006optical}) between the two frames are extracted and fed into a 2D CNN, reducing the computation cost while retaining most of the features~\cite{van2019capsulenet,zhang2022short,hochreiter1997long}.
In the other category, a sequence of dynamic feature maps between every two adjacent frames are extracted and learned by a time series network or a 3D CNN (\eg, ELRCN~\cite{khor2018enriched}, STRCN-A~\cite{xia2019spatiotemporal} and 3DFlowNet~\cite{li2019micro}), taking the spatio-temporal features from the whole sequence as input.
The entire image sequence can also be compressed into a single dynamic feature map (\eg, dynamic imaging~\cite{bilen2016dynamic,verma2019learnet}) and then processed (\eg, LEARNet~\cite{verma2019learnet} and AffectiveNet~\cite{verma2020affectivenet}), maintaining both high-level and micro-level information.
However, the LOSO validation accuracy of networks based on image sequences is generally lower than those using image pairs as input, probably because the information redundancy makes it difficult to focus on the most important features. Following the mainstream methods, our FRL-DGT model also takes an onset-apex frame pair as input, but we still extend it on the whole sequence to validate its superiority.

{\bf Dynamic Features} are good at capturing subtle changes existing in the frame sequences. As a common type, optical flow and its variants (\eg, Bi-WOOF~\cite{liong2018less} and MDMO~\cite{liu2018sparse}) can estimate the direction and magnitude of the displacement between two frames and extract the inter-frame motion information. Another useful type is dynamic image~\cite{bilen2016dynamic}, which is a single RGB image obtained by compressing both the spatial information and temporal dynamic features from the image sequence. To some extent, both these two kinds of dynamic features have been successfully applied to ME recognition~\cite{liong2018less,verma2019learnet}. Instead of using the common dynamic features, we design an adaptive DGM into the end-to-end pipeline so that we are able to extract the more targeted dynamic features for ME classification.

{\bf Visual Transformers}~\cite{vaswani2017attention, dosovitskiy2020image, Dong:MM2021, bertasius2021space} have evolved rapidly with powerful variants developed for image and video classification. The input frames are split into evenly divided fixed-size patches, which are linearly projected into tokens and fed into a Transformer encoder. Obviously, such division may divide the key parts into different patches, and is independent of image content. Inspired by Transformer iN Transformer (TNT)~\cite{han2021transformer} and Swin Transformer~\cite{liu2021swin}, we take AU regions as input to our Transformer Fusion to extract local features and learn global information.

\section{Proposed Method}
We propose a novel end-to-end ME recognition network, named FRL-DGT, as shown in Figure~\ref{fig:Overall_structure}. It takes an onset-apex image pair from a frame sequence as input, and generates displacement features between them through the DGM trained with self-supervision (Section~\ref{sec:DGM}). The displacement is concatenated with \textcolor{black}{the corresponding onset-apex pair}, cropped according to the AU regions and full-face region, and then fed into the Transformer Fusion (Section~\ref{sec:FuTrans}) to obtain strong feature representation for ME classification.

\subsection{Displacement Generation Module with Self-supervised Learning}\label{sec:DGM}

The Displacement Generation Module (DGM) here is designed to extract the adaptive dynamic features for the specific ME task. It takes the onset-apex image pair as input and outputs a pixel displacement feature map $D$ between the two frames. The structure of DGM follows an encoder-decoder style, first downsampling the high-resolution input images to obtain low-dimensional dynamic features, and then upsampling them to obtain displacements between image pairs, as shown in Figure~\ref{fig:Overall_structure}. The basic block in DGM is a stack of convolutional layers, batch normalization layers, and nonlinear activation layers. Note that we normalize the displacements before output, increasing the intensity of minor expressions and decreasing the intensity of major ones, which plays a role of adaptive expression adjustment.

\begin{figure}[ht]
\captionsetup[subfloat]{justification=centering, singlelinecheck=false, labelformat=empty} 
\centering
  \subfloat[]{\includegraphics[height=0.09\textwidth]{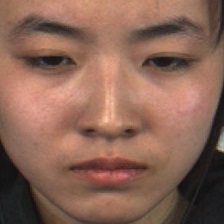}}
  \subfloat[]{\includegraphics[height=0.09\textwidth]{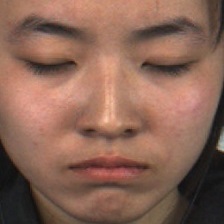}}
  \subfloat[]{\includegraphics[height=0.09\textwidth]{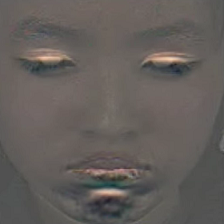}}
  \subfloat[]{\includegraphics[height=0.09\textwidth]{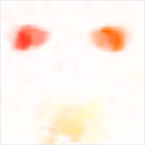}}
  \subfloat[]{\includegraphics[height=0.09\textwidth]{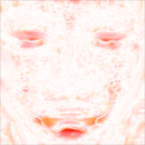}} 

  \vspace{-12pt}
  \subfloat[]{\includegraphics[height=0.09\textwidth]{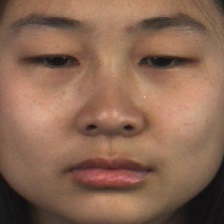}}
  \subfloat[]{\includegraphics[height=0.09\textwidth]{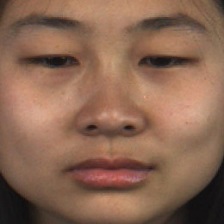}}
  \subfloat[]{\includegraphics[height=0.09\textwidth]{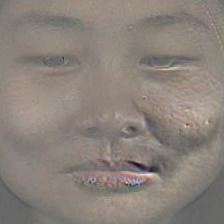}}
  \subfloat[]{\includegraphics[height=0.09\textwidth]{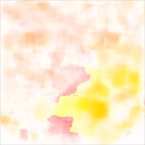}}
  \subfloat[]{\includegraphics[height=0.09\textwidth]{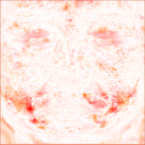}}

  \vspace{-12pt}
  \subfloat[(a)]{\includegraphics[height=0.09\textwidth]{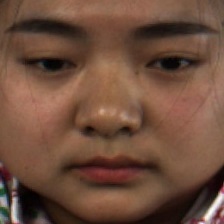}}
  \subfloat[(b)]{\includegraphics[height=0.09\textwidth]{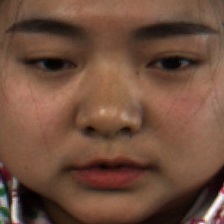}}
  \subfloat[(c)]{\includegraphics[height=0.09\textwidth]{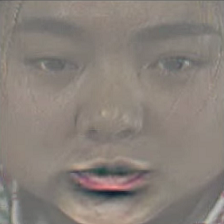}}
  \subfloat[(d)]{\includegraphics[height=0.09\textwidth]{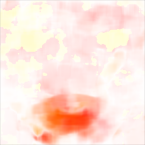}}
  \subfloat[(e)]{\includegraphics[height=0.09\textwidth]{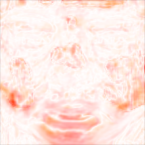}}
  \vspace{-0.3cm}		
   \caption{Visualization of the generated displacement on three ME categories: surprise, positive, and negative (from top to bottom). On each row from left to right are (a) Onset, (b) Apex, (c) Dynamic image, (d) Optical flow, and (e) Our displacement, respectively. (Images from CASME II \copyright{Xiaolan Fu}
   )}
  \label{fig:vis_dgm}
\vspace{-0.35cm}
\end{figure}

Similar to optical flow features, the displacement values represent the relative pixel position shifts in x and y directions from onset frame to apex frame. To limit the range of movable pixel positions and make the model easier to learn, we multiply the output displacement in the range [-1, 1] by a scaling factor $\alpha$. Three loss components are set for the output displacement: reconstruction loss $\mathcal{L}_{rec}$, normalization loss $\mathcal{L}_{nm}$, and smoothing loss $\mathcal{L}_{sm}$.

Let $F_{onset}$ be the onset frame, $F_{apex}$ be the original apex frame, $D_{x,y}$ be the displacement value at (x, y), and $\mathbb{G}_s(F_{onset}, D)$ represents the approximate apex frame obtained by bilinear sampling the onset frame according to the generated displacement $D$. That is, bilinear sampling moves the pixel at location (x, y) in the onset frame to (x+$D_{x,y}^x$, y+$D_{x,y}^y$) of the approximate apex frame. Then the displacement related loss $\mathcal{L}_{DGM}$ is calculated by:
\begin{equation}
    \mathcal{L}_{rec} = |\mathbb{G}_s(F_{onset}, D)-F_{apex}|,
    \label{eq:1}
\end{equation}
\begin{equation}
    \mathcal{L}_{nm} = \frac{1}{w\times h}\sum_{x=0}^{w-1} \sum_{y=0}^{h-1} |D_{x,y}|,
    \label{eq:2}
\end{equation}
\begin{gather}
    \mathcal{L}_{sm} = \frac{\sum_{y=0}^{h-1} \sum_{x=1}^{w-1} |D_{x,y}-D_{x\!-\!1,y}|}{h\times(w-1)} + \notag \\ 
    \frac{\sum_{x=0}^{w-1} \sum_{y=1}^{h-1} |D_{x,y}-D_{x,y\!-\!1}|}{w\times(h-1)},
    \label{eq:3}
\end{gather}
\begin{equation}
    \mathcal{L}_{DGM} = \lambda_{rec}\mathcal{L}_{rec}+\lambda_{nm}\mathcal{L}_{nm}+\lambda_{sm}\mathcal{L}_{sm},
    \label{eq:4}
\end{equation}
where ($h$, $w$) is the size of images, $\lambda_{rec}$, $\lambda_{nm}$ and $\lambda_{sm}$ are different weights assigned to each loss component.


Note that only $\mathcal{L}_{DGM}$ is not enough to generate the dynamic features specific for ME recognition, we have to combine it to classification loss $\mathcal{L}_{cls}$ together into the end-to-end training. To avoid the underfitting issue due to the limited ME data samples, we resort to a self-supervised learning strategy by sampling sufficient number of image pairs with the $\mathcal{L}_{DGM}$ loss applied only. It is worthy mentioning that self-supervision here is very critical due to the number of training images in the field of ME recognition. 

To make it easy to understand our DGM module, we visualize the generated displacements in Figure~\ref{fig:vis_dgm}. Our generated displacement is more targeted to AU regions and more sensitive to capture the subtle changes of related MEs.

\subsection{Transformer Fusion on \textcolor{black}{Onset-Apex Pair} and Extracted Displacement}
\label{sec:FuTrans}
\subsubsection{AU Regions} 
\begin{figure}[ht]
\captionsetup[subfloat]{justification=centering, singlelinecheck=false, labelformat=empty} 
\vspace{-0.25cm}	
\centering
  \subfloat[(a)]{\includegraphics[height=0.1\textwidth]{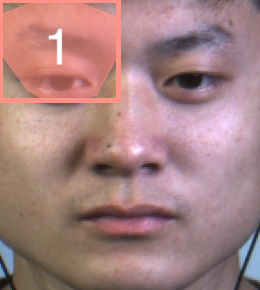}}
  \subfloat[(b)]{\includegraphics[height=0.1\textwidth]{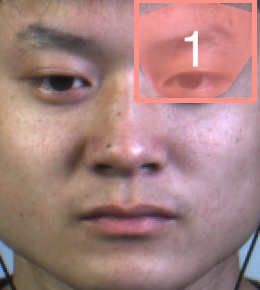}} 
  \subfloat[(c)]{\includegraphics[height=0.1\textwidth]{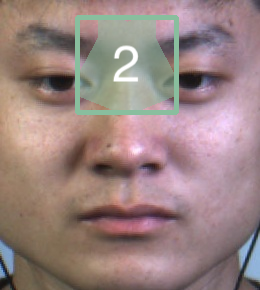}} 
  \subfloat[(d)]{\includegraphics[height=0.1\textwidth]{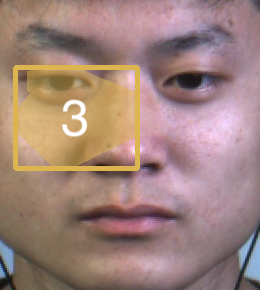}} 
  \subfloat[(e)]{\includegraphics[height=0.1\textwidth]{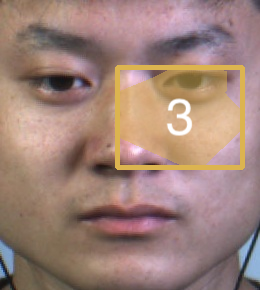}}
  \vspace{0.0cm}
  \subfloat[(f)]{\includegraphics[height=0.1\textwidth]{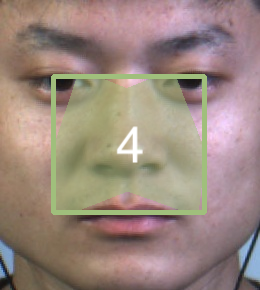}}
  \subfloat[(g)]{\includegraphics[height=0.1\textwidth]{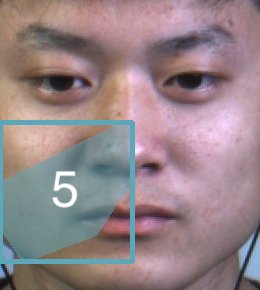}}
  \subfloat[(h)]{\includegraphics[height=0.1\textwidth]{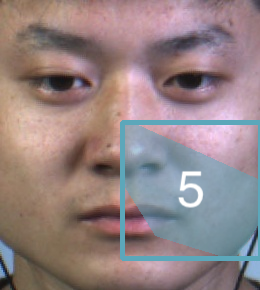}}
  \subfloat[(i)]{\includegraphics[height=0.1\textwidth]{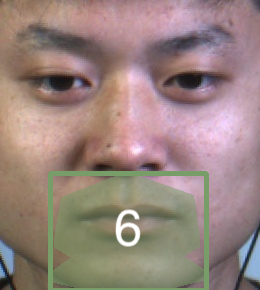}}
  {\includegraphics[height=0.09\textwidth]{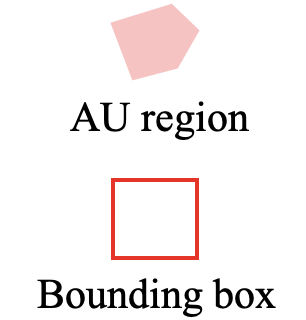}}
  \vspace{-0.3cm}		
   \caption{Visualization of 9 AUs as input of Transformer Fusion module. (a) AU region \#1 left, (b) AU region \#1 right, (c) AU region \#2, (d) AU region \#3 left, (e) AU region \#3 right, (f) AU region \#4, (g) AU region \#5 left, (h) AU region \#5 right, and (i) AU region \#6. (Images from CASME II \copyright{Xiaolan Fu})}
  \label{fig:AU_regions}
\vspace{-0.55cm}
\end{figure}

AU regions are divided with reference to the rules in~\cite{ma2019r}, {\em i.e.}, 68 landmark points are obtained using the
dlib package, and the face is divided into 43 basic region-of-interest (RoIs). As shown in Figure~\ref{fig:AU_regions}, 9 AUs are selected and each AU corresponds to multiple RoIs based on the relationship between MEs and facial muscle movements~\cite{eckman1978facial,ekman1971constants}.

In order to make the size of each bounding box appropriate, AU regions \#1, \#3, and \#5 (marked in red, yellow, and blue, respectively in Figure~\ref{fig:AU_regions}) are all divided into left and right parts of each. AU regions \#1 and \#2 are mainly concerned with changes above the eyes, such as eyebrow lifting and frowning. AU regions \#3 and \#4 are responsible for changes in the middle of the face, around the nose and lower eyelids. While the changes at the mouth, chin and cheeks are focused on by AU regions \#5 and \#6.

\subsubsection{Feature Fusion Modules with Transformer} 
The three Transformer-based feature fusion modules (local, global and full-face fusions) perform learning and fusion of embedding vectors in a hierarchical manner based on $K$ cropped AU regions and the full-face region. 
\textcolor{black}{The target AU boxes can be set to different resolutions, and the $i$-th AU box corresponds to size ($H_i$, $W_i$).} 
The number of channels $M$ is the same for all AU regions (gray or colorful), which is equal to that of the input feature map.

Similar to the method in Vision Transformer (ViT)~\cite{dosovitskiy2020image} with patch size $P\times P$, we divide each AU $x_r\in \mathbb{R}^{H_i \times W_i \times M}$ into a sequence of image patches $x_p\in \mathbb{R}^{N \times P \times P \times M}$, where $N=H_iW_i/P^2$  is the resulting number of patches. The linear projection maps each patch to a $C$-dimensional embedding vector ({\em i.e.}, vector length is $C$) without using the class token. Then both local and global feature fusions are applied to the selected AUs to obtain strong feature representation for ME classification, as shown in Figure~\ref{fig:Overall_structure}.

\begin{figure}[t]
  \centering
   \subfloat[Fusion after attention]{
   \hspace{-0.25cm}
   \includegraphics[width=1.0\linewidth]{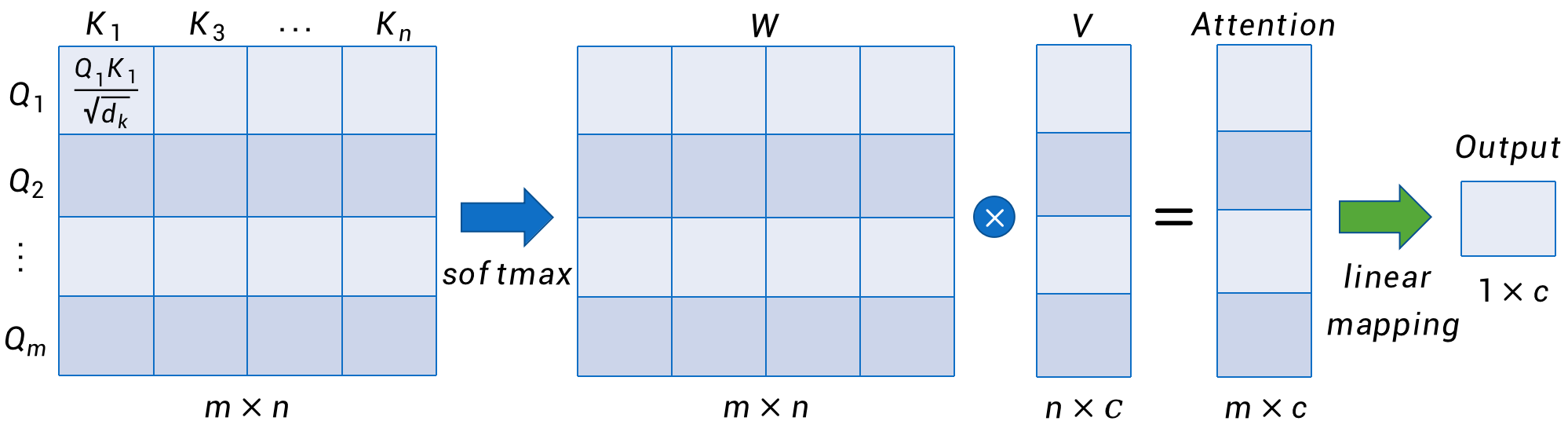}}
   \newline
   \subfloat[Fusion before attention]{\includegraphics[width=1.0\linewidth]{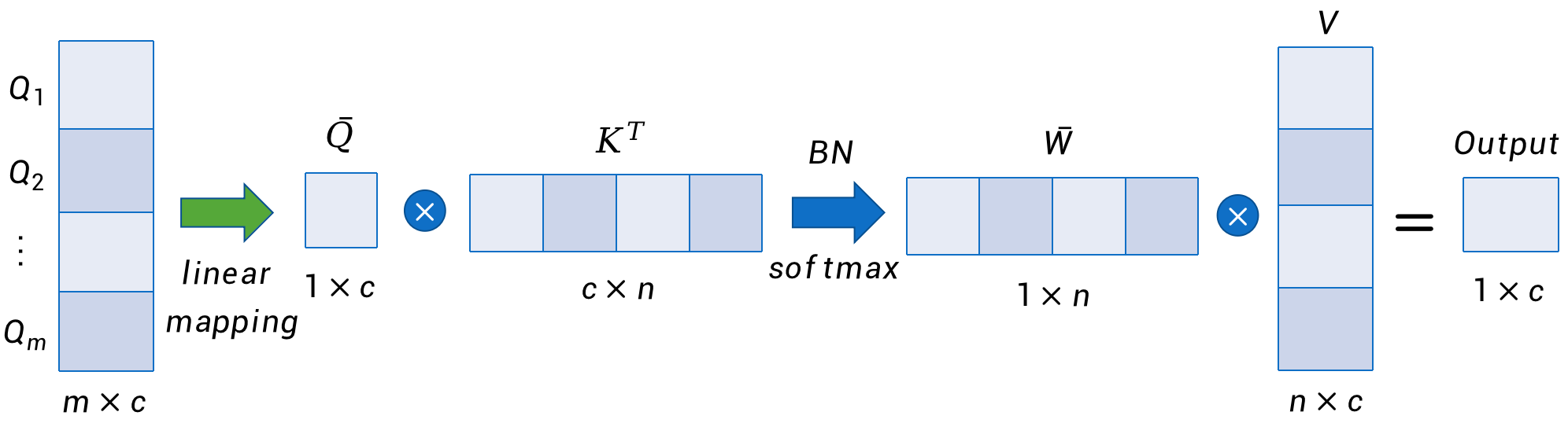}}
   \vspace{-0.2cm}
   \caption{Comparison of attention and fusion mechanisms. Different from the original attentions with (a) linear fusion, our proposed fusion layer (b) obtains the linear mapped queries first, which has the advantages of reducing noise and simplifying computation.
   }
   \label{fig:FusionNew}
   \vspace{-0.40cm}
\end{figure}

{\bf Fusion with Attention}. Each embedding vector is dot-multiplied with the weight matrices to obtain the query, key and value, which are all $C$-dimensional vectors. For merging, the vectors of $N \times C$ output from the attention layer can be mapped to $C$ dimensions 
by a linear transformation, as shown in Figure~\ref{fig:FusionNew} (a). In contrast, for the fusion layer in Figure~\ref{fig:FusionNew} (b), we perform linear mapping to the queries of each key before dot product to remove the influence of noise, then pass it through a Batch Normalization layer and a Softmax function to obtain a probability vector that represents the relative importance of it in overall sequence, and the probabilities are used to weight the sum of all values.

The fusion layer adopts a multi-head mechanism that allows the model to learn different importance distributions of embedding vectors in multiple subspaces. 
For each head, we set $c=C/h$, where $h$ is the number of attention heads. Packing the $k$-th head  part ($h$ parts in total) of the embedding vector of queries, keys and values together into matrices $Q \in \mathbb{R}^{m \times c}$, $K \in \mathbb{R}^{n \times c}$ and $V \in \mathbb{R}^{n \times c}$ respectively \textcolor{black}{(Note that $m=n$ in our case)}, the fused embedding vector of the $k$-th head is obtained by:
\begin{equation}
\overline{W}_{raw}= Lin(\left[Q_1, Q_2, ..., Q_m \right]) \cdot K^T,
    \label{eq:7}
\end{equation}
\begin{equation}
    head_k = Softmax(BN(\overline{W}_{raw})) \cdot V,
    \label{eq:8}
\end{equation}
\textcolor{black}{where $Lin$ performs the linear transformation of $Q$}, and $BN$ stands for the batch normalization layers. 
The final fusion result is obtained by concatenating the fused embedding vector $F$ from $h$ heads:
\begin{equation}
    F=Concat(head_1,\dots,head_h),
    \label{eq:9}
\end{equation}

Note that when all weights of linear mappings are equal, it corresponds to an averaging operation. The fusion before attention mechanism is used in the fusion layers of local, global and full-face modules in Transformer Fusion.

{\bf Local Fusion Module}. The first level contains $K$ local modules, and different AUs are processed by different local modules respectively. The $i$-th local module performs feature extraction on the $N$ patches of the $i$-th AU and fuses them into a local embedding vector that contains the spatial (\textcolor{black}{onset-apex pair}) and dynamic (displacement) features.

{\bf Global Fusion Module}. The second level contains a spatial module that learns and fuses the $K$ local embedding vectors outputted from the previous level to obtain the expression information contained in each frame, represented by a global embedding vector.

{\bf Full-face Fusion Module}. In addition to the hierarchical features of AUs, we perform attention learning and spatial feature fusion on entire face images to obtain full-face features as auxiliary classification information. Note that each module in local, global and full-face feature fusions consists of an Attn block for feature learning and a fusion layer for information synthesis. Attention Learning (Attn) block is a stack of multiple Transformer layers, including Layer Normalization, Multi-head Attention and Multi-layer Perceptron. The resulting embedding vector is fed into the subsequent fully connected layers for ME classification.


{\bf Discussion}: Compared to CNN networks, basic Transformer structures lack translation equivariance and locality, thus generalizing poorly when the amount of training data is insufficient. To overcome the weakness, we calibrate the face to a fixed position and size before performing recognition, so that we do not need to pay much attention to rotation and translation invariance. Then we build the Transformer architecture in the form of multi-level fusion, which allows better focus on local features. Moreover, linear fusion before attention can suppress noise and thus is beneficial to error elimination, meanwhile, our method reduces the amount of calculation by early using a vector instead of a matrix.



\renewcommand\arraystretch{1.1}
\renewcommand{\multirowsetup}{\centering}
\begin{table*}[h]
  \vspace{-0.20cm}
  \centering
  \small
  \resizebox{0.8\textwidth}{!}{\begin{tabular}{l|c|c|cccccccc}
    \toprule
    \multirow{2}{*}{Method} & \multirow{2}{*}{Year} & \multirow{2}{*}{Type} &
    \multicolumn{2}{c}{Full} & \multicolumn{2}{c}{SMIC Part} & 
    \multicolumn{2}{c}{SAMM Part} & \multicolumn{2}{c}{CASME II Part} \\
    \cline{4-11}
    & & & UF1 & UAR & UF1 & UAR & UF1 & UAR & UF1 & UAR \\
    \midrule
    LBP-TOP~\cite{yan2014casme} & 2014 & Hand-Crafted & 0.588 & 0.579 & 0.200 & 0.528 & 0.395 & 0.410 & 0.703 & 0.743 \\
    Bi-WOOF~\cite{liong2018less} & 2018 & Hand-Crafted & 0.630 & 0.623 & 0.573 & 0.583 & 0.521 & 0.514 & 0.781 & 0.803 \\
    CapsuleNet~\cite{van2019capsulenet} & 2019 & Deep-Learning & 0.652 & 0.651 & 0.582 & 0.588 & 0.621 & 0.599 & 0.707 & 0.702 \\
    STSTNet~\cite{liong2019shallow} & 2019 & Deep-Learning & 0.735 & 0.761 & 0.680 & 0.701 & 0.659 & 0.681 & 0.838 & 0.869 \\
    RCN-A~\cite{xia2020revealing} & 2020 & Deep-Learning & 0.743 & 0.719 & 0.633 & 0.644 & \textcolor{blue}{0.760} & 0.672 & 0.851 & 0.812 \\
    GEME~\cite{nie2021} & 2021 & Deep-Learning & 0.740 & 0.750 & 0.629 & 0.657 & 0.687 & 0.654 & 0.840 & 0.851 \\
    MERSiamC3D~\cite{zhao2021two} & 2021 & Deep-Learning & \textcolor{blue}{0.807} & \textcolor{blue}{0.799} & \textcolor{blue}{0.736} & \textcolor{red}{0.760} & 0.748 & \textcolor{blue}{0.728} & 0.882 & 0.876 \\
    FeatRef~\cite{zhou2022} & 2022 & Deep-Learning & 0.784 & 0.783 & 0.701 & 0.708 & 0.737 & 0.716 & \textcolor{blue}{0.892} & \textcolor{blue}{0.887} \\
    \midrule
    FRL-DGT & 2022 & Deep-Learning & \textcolor{red}{0.812} & \textcolor{red}{0.811} & \textcolor{red}{0.743} & \textcolor{blue}{0.749} & \textcolor{red}{0.772} & \textcolor{red}{0.758} & \textcolor{red}{0.919} & \textcolor{red}{0.903} \\
    \midrule
    EMRNet~\cite{liu2019neural}* & 2019 & Deep-Learning & 0.789 & 0.782 & \underline{0.746} & \underline{0.753} & \underline{0.775} & 0.715 & 0.829 & 0.821 \\
    FGRL-AUF~\cite{lei2021micro}* & 2021 & Deep-Learning & 0.791 & 0.793 & 0.719 & 0.722 & \underline{0.775} & \underline{0.789} & 0.880 & 0.871 \\
    ME-PLAN~\cite{ME-PLAN2022}* & 2022 & Deep-Learning & 0.772 & 0.786 & 0.713 & 0.726 & 0.716 & 0.742 & 0.863 & 0.878 \\
    \bottomrule
  \end{tabular}}
  \vspace{-0.15cm}
  \caption{Performance comparison of the SOTA methods and our proposed FRL-DGT in terms of UF1 and UAR. The best and second best results are marked in red and \textcolor{black}{blue} colors, respectively. Methods with * use different datasets, and they have underlined higher scores.}
  \label{tab:comparative experiments for FRL-DGT}
  \vspace{-0.30cm}
\end{table*}

\subsection{Implementation Details}
Each of the selected $K=9$ AUs is with a size of (90, 90), the patch size is (18, 18), and the number of patches is $N=25$. The dimension $C$ of embedding vectors is 256, while both the attention and fusion layers use $h=8$ heads. In the formulas of DGM, we take $\lambda_{rec}=10$, $\lambda_{nm}=1$, $\lambda_{sm}=0.2$, and displacement scaling factor $\alpha=0.2$.
The batch size is 32, and the optimizer is Adam with an initial learning rate of 0.002 for DGM, and the Adam with cosine annealing strategy for Transformer Fusion. \textcolor{black}{The gradient in DGM back-propagated from the classification loss is scaled by $10^{-6}$ to prevent it from being dominant.}

Note that to increase the amount of image pairs for training and improve the sensitivity of our network to the subtle expression changes, we use MagNet~\cite{Oh2018Learning} to augment the datasets as in \cite{lei2021micro}, and also perform a randomization operation: select a frame before or after the original apex frame randomly when loading the data for training.



\section{Experiments}


\subsection{Datasets and Metrics} 
Since the Composite Database Evaluation (CDE) is commonly used for evaluation and comparison in 
the field of ME recognition because of the small amount of collected samples, we use 3 ME datasets CASME II~\cite{yan2014casme}, SAMM~\cite{davison2016samm} and SMIC~\cite{li2013spontaneous,Pfister2011Recognising} for composite training\footnote{The three datasets were received and exclusively accessed by the author Zhijun Zhai and Jianhui Zhao for purely academic research only. The author Zhijun Zhai produced the experimental results
in this paper. Meta did not have access to the datasets as part of this research.}. And we adopt the same way as in MEGC2019 Challenge~\cite{see2019megc} to unify different category settings across datasets, mapping them to 3 general classes: \textbf{Negative}\{ ``Repression", ``Anger", ``Contempt", ``Disgust", ``Fear",  ``Sadness" \}, \textbf{Positive}\{``Happiness"\}, and \textbf{Surprise}\{``Surprise"\}.
Unrelated or undefined emotion categories such as ``Others" are omitted to reduce confusion for model training. 

The 3 ME datasets have 442 image sequences from 68 subjects, 25,469 images in total with 58 averaged frames per sequence. To clarify, the onset (starting time of ME), apex (time with the highest intensity of ME), and offset (ending time of ME) frames are already labeled and provided in the benchmark datasets SAMM and CASME II. 
\textcolor{black}{We follow~\cite{liu2019neural} to obtain the apex frames of image sequences in SMIC,}
which are not officially labeled. For real scenarios, there are a lot of proven methods to locate the onset, apex, and offset frames from a video~\cite{ben2021video,li2017towards}.
\textcolor{black}{After obtaining the key frames, we perform face calibration with dlib package to handle image-plane rotation and translation.}



Regarding the evaluation metrics, we follow~\cite{liu2019neural,van2019capsulenet}, taking Unweighted F1-score (UF1) and Average Recall (UAR) with leave-one-subject-out (LOSO) cross validation to evaluate the ME recognition performance.

\begin{figure}[ht!]
  \centering
    \subfloat[EMRNet (From left to right: Full, SMIC, SAMM, and CASME II)]{
      \includegraphics[height=0.11\textwidth]{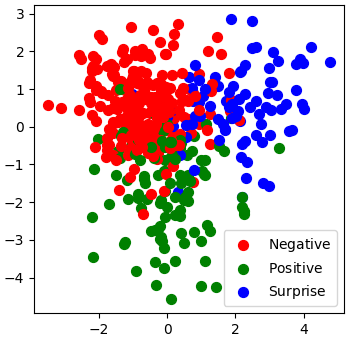}
     \hspace{-0.15cm}
     \includegraphics[height=0.11\textwidth]{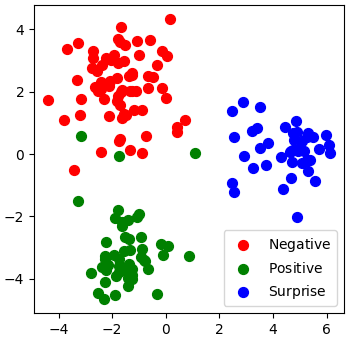}
     \hspace{-0.15cm}
     \includegraphics[height=0.11\textwidth]{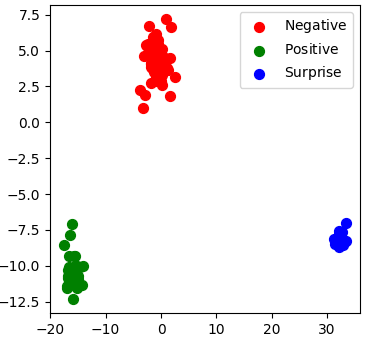} 
     \hspace{-0.17cm}
     \includegraphics[height=0.11\textwidth]{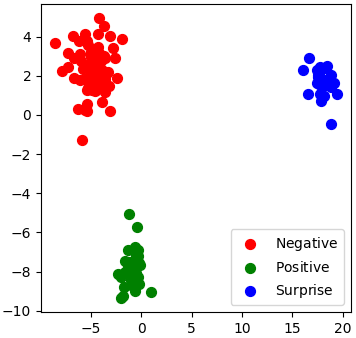} }
    \newline
    \subfloat[FGRL-AUF (From left to right: Full, SMIC, SAMM, and CASME II)]{
      \includegraphics[height=0.11\textwidth]{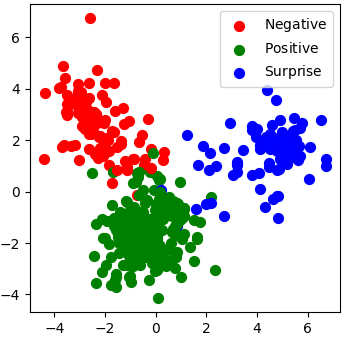}
     \includegraphics[height=0.11\textwidth]{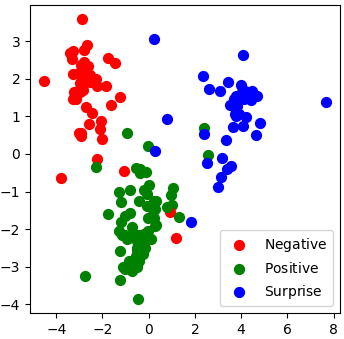}
     \includegraphics[height=0.11\textwidth]{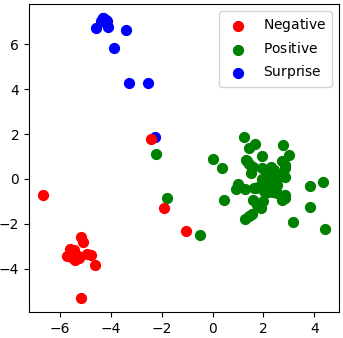} 
     \includegraphics[height=0.11\textwidth]{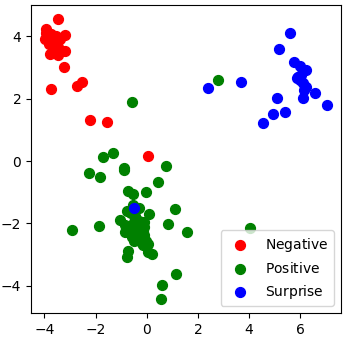} }
    \newline
    \subfloat[FRL-DGT (From left to right: Full, SMIC, SAMM, and CASME II)]{
      \hspace{-0.05cm}
      \includegraphics[height=0.11\textwidth]{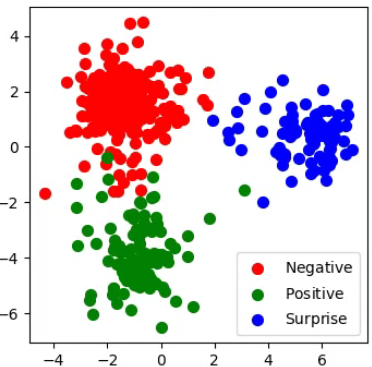}
      \includegraphics[height=0.11\textwidth]{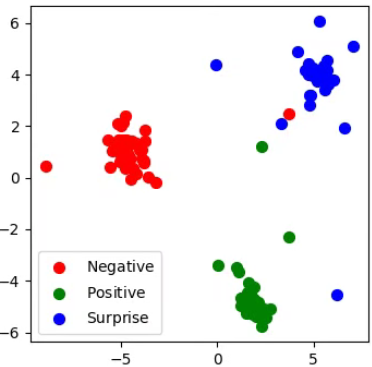}
      \includegraphics[height=0.11\textwidth]{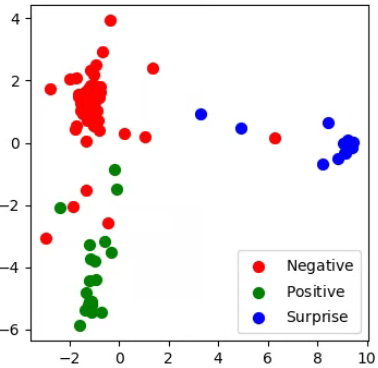}
      \includegraphics[height=0.11\textwidth]{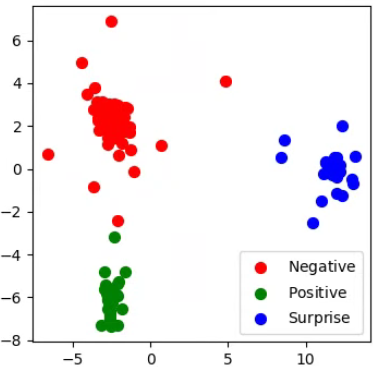} }\\
   \vspace{-0.15cm}
   \caption{The feature distributions of EMRNet, FGRL-AUF and our proposed FRL-DGT on the evaluation datasets.}
   \label{fig:Viz_feature}
   \vspace{-0.30cm}
\end{figure}

\subsection{Comparison to State-of-the-art Methods}

The comparative SOTA methods include representative works of two main ME recognition categories and mainstream approaches based on deep learning in recent years. As shown in Table~\ref{tab:comparative experiments for FRL-DGT}, both UF1 and UAR of our framework FRL-DGT are higher than 0.810. FRL-DGT improves by 0.62\% and 1.50\% on UF1 and UAR over MERSiamC3D, the second best deep learning based ME classifier.

\begin{table*}[ht!]
  \vspace{-0.20cm}
  \centering
  \resizebox{\linewidth}{!}{\begin{tabular}{c|cccccc|cccccccc}
    \toprule
    \multirow{2}{0.06\textwidth}{Method} &\multirow{2}{0.04\textwidth}{DGM} &\multirow{2}{0.06\textwidth}{AU Regions} &\multirow{2}{0.08\textwidth}{Full-face Fusion} &\multirow{2}{0.05\textwidth}{Global Fusion} & \multirow{2}{0.05\textwidth}{Local Fusion} &\multirow{2}{0.12\textwidth}{Fu-B-Attn} &
    \multicolumn{2}{c}{Full} & \multicolumn{2}{c}{SMIC Part} & 
    \multicolumn{2}{c}{SAMM Part} & \multicolumn{2}{c}{CASME II Part} \\
    \cline{8-15}
    & & & & & & & UF1 & UAR & UF1 & UAR & UF1 & UAR & UF1 & UAR \\
    \midrule
    M0 & $\rightarrow$OpticalFlow & $\checkmark$ & $\checkmark$ & $\checkmark$ & $\checkmark$ & $\checkmark$ & 0.741 & 0.718 & 0.671 & 0.662 & 0.695 & 0.662 & 0.846 & 0.834 \\
    M1 & $\rightarrow$OF+NORM & $\checkmark$ & $\checkmark$ & $\checkmark$ & $\checkmark$ & $\checkmark$ & 0.758 & 0.739 & 0.671 & 0.667 & \textcolor{red}{0.778} & 0.730 & 0.869 & 0.831 \\
    M2 & $\rightarrow$DynamicImage & $\checkmark$ & $\checkmark$ & $\checkmark$ & $\checkmark$ & $\checkmark$ & 0.739 & 0.720 & 0.684 & 0.679 & 0.762 & \textcolor{blue}{0.745} & 0.759 & 0.716 \\
    \textcolor{black}{M3} & w/o self-supervise & $\checkmark$ & $\checkmark$ & $\checkmark$ & $\checkmark$ & $\checkmark$ & 0.778 & 0.777 & 0.707 & 0.718 & 0.697 & 0.677 & \textcolor{blue}{0.914} & \textcolor{blue}{0.889} \\
    \midrule
    \textcolor{black}{M4} & $\checkmark$  & $\checkmark$ & $\checkmark$ & $\checkmark$ & $\checkmark$ & $\rightarrow$Fu-A-Attn & \textcolor{blue}{0.797} & \textcolor{blue}{0.792} & \textcolor{red}{0.746} & \textcolor{blue}{0.746} & 0.734 & 0.719 & 0.898 & 0.885 \\
    \midrule
    \textcolor{black}{M5} & $\checkmark$ & $\rightarrow$3x3 image patches & $\checkmark$ & $\checkmark$ & $\checkmark$ & $\checkmark$ & 0.765 & 0.765 & 0.665 & 0.673 & 0.754 & 0.734 & 0.894 & 0.876 \\
    \textcolor{black}{M6} & $\checkmark$  & $\checkmark$ & $\times$ & $\checkmark$ & $\checkmark$ & $\checkmark$ & 0.773 & 0.774 & 0.689 & 0.698 & 0.758 & 0.704 & 0.876 & 0.881 \\
    \textcolor{black}{M7} & $\checkmark$  & $\checkmark$ & $\checkmark$ & $\checkmark$ & $\times$ & $\checkmark$ & 0.781 & 0.765 & 0.741 & 0.745 & 0.725 & 0.672 & 0.848 & 0.838 \\
    \textcolor{black}{M8} & $\checkmark$  & $\checkmark$ & $\checkmark$ & $\times$ & $\checkmark$ & $\checkmark$ & 0.782 & 0.773 & 0.701 & 0.706 & 0.711 & 0.671 & 0.904 & 0.886 \\
    \textcolor{black}{M9} & $\checkmark$  & $\checkmark$ & $\checkmark$ & $\checkmark$ & $\checkmark$ & $\checkmark$ & \textcolor{red}{0.812} & \textcolor{red}{0.811} & \textcolor{blue}{0.743} & \textcolor{red}{0.749} & \textcolor{blue}{0.772} & \textcolor{red}{0.758} & \textcolor{red}{0.919} & \textcolor{red}{0.903} \\
    \bottomrule
  \end{tabular}}
  \vspace{-0.15cm}
  \caption{Ablation study of our proposed network. ``$\rightarrow$X" indicates replacing the corresponding component with X. $\checkmark$ and $\times$ represent yes or no, respectively. The best and second best results are marked in red and \textcolor{black}{blue} colors, respectively.}
  \label{tab:ablation experiments}
  \vspace{-0.25cm}
\end{table*}

\begin{figure*}[t]
  \centering
   \includegraphics[width=1.0\linewidth]{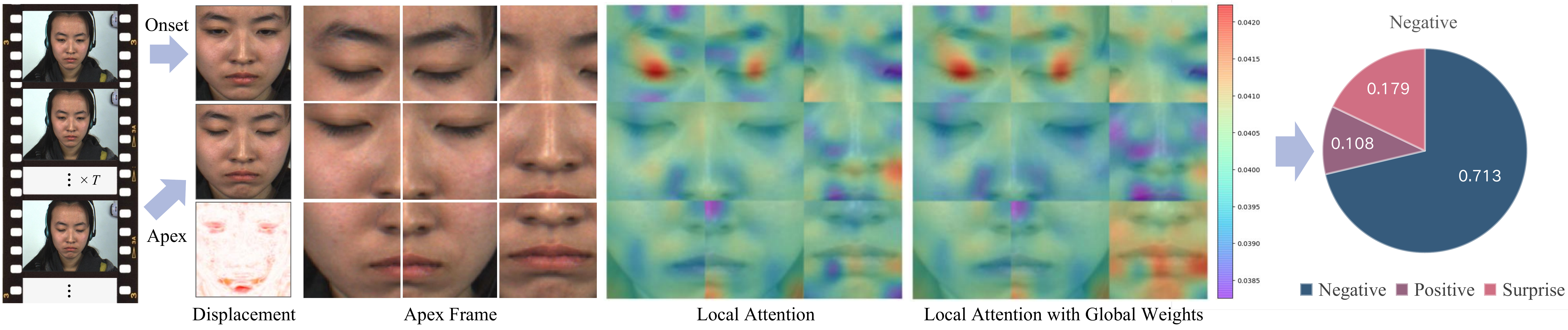}
   \vspace{-0.75cm}
   \caption{Visualization of the weights in fusion layers. (Images from CASME II \copyright{Xiaolan Fu})}
   \label{fig:Viz_fusion}
   \vspace{-0.15cm}
\end{figure*}

There are other existing efficient methods \cite{yu2020ice,gupta2021merastc}, \eg, EMRNet\cite{liu2019neural} introduces CK+ dataset to implement domain adaptation, FGRL-AUF\cite{lei2021micro} uses only CASME II and SAMM datasets for the annotated AUs, and ME-PLAN\cite{ME-PLAN2022} constructs a pre-training by combining macro-expression samples in CK+, Oulu-CASIA and DFEW datasets. They may have higher scores on certain dataset, but their full scores are still less than our FRL-DGT.
As shown in Figure~\ref{fig:Viz_feature}, the distribution of extracted features from our FRL-DGT is more separated than those of EMRNet and FGRL-AUF, resulting in better classification results. Taken together, our network has excellent results in the same type of methods and is highly competitive among them.


\subsection{Ablation Study}


\textbf{Displacement Generating Module}. The superiority of DGM over the two conventional methods is demonstrated in Table~\ref{tab:ablation experiments}, where DynamicImage stands for dynamic imaging method and OpticalFlow stands for optical flow method. The ME performance by replacing DGM with optical flow or dynamic imaging is lower than that of FRL-DGT, which is an end-to-end network combining DGM and Transformer Fusion. 
\textcolor{black}{We also replace DGM with the normalized OpticalFlow (OF+NORM), except for the boost of accuracy on SAMM by NORM, the overall performance is still inferior to that of DGM.}
The dynamic features obtained by the three approaches are visualized in Figure~\ref{fig:vis_dgm}. It can be seen that all the three dynamic features can highlight the changing regions, but the displacement generated by DGM can be automatically adjusted according to the classification loss, obtaining additional hidden information. Thus, the following Transformer Fusion module can carry on more objective learning and get more reliable classification results. From M3 and M9 in Table~\ref{tab:ablation experiments}, we can also find that self-supervised learning is very useful for DGM, helps improve the full UF1 score significantly from 0.778 to 0.812 by 4.37$\%$, and the UAR score significantly from 0.777 to 0.811 by 4.38$\%$.

\textbf{Transformer Fusion}. The fusion layer with linear fusion before attention (Fu-B-Attn) of M9 is replaced with a normal linear fusion after attention (Fu-A-Attn) of M4.
Comparison between M4 and M9 in Table~\ref{tab:ablation experiments} demonstrates that our Fu-B-Attn merges embedding vectors in a more effective way. For runtime, Fu-A-Attn takes 50.8ms while Fu-B-Attn only needs 47.6ms to run all fusions, 
which proves that our Fu-B-Attn can simplify computation.

Based on the results of M6, M7, and M8, there are contributions from three fusion modules in Transformer Fusion, i.e., local fusion module, global fusion module, and full-face fusion module, respectively.
Figure~\ref{fig:Viz_fusion} plots the original weight distribution in local fusion layer with each AU region and the weight distribution after global fusion, which shows that the global fusion further adjusts the attention to make it more focused on the eyes and mouth regions. Note that we compress the global fusion weights while maintaining the comparison order to facilitate visualization.

\begin{figure*}[ht]
  \centering
    \subfloat[4 classes]{
    \hspace{-0.25cm}
      \includegraphics[height=0.22\textwidth]{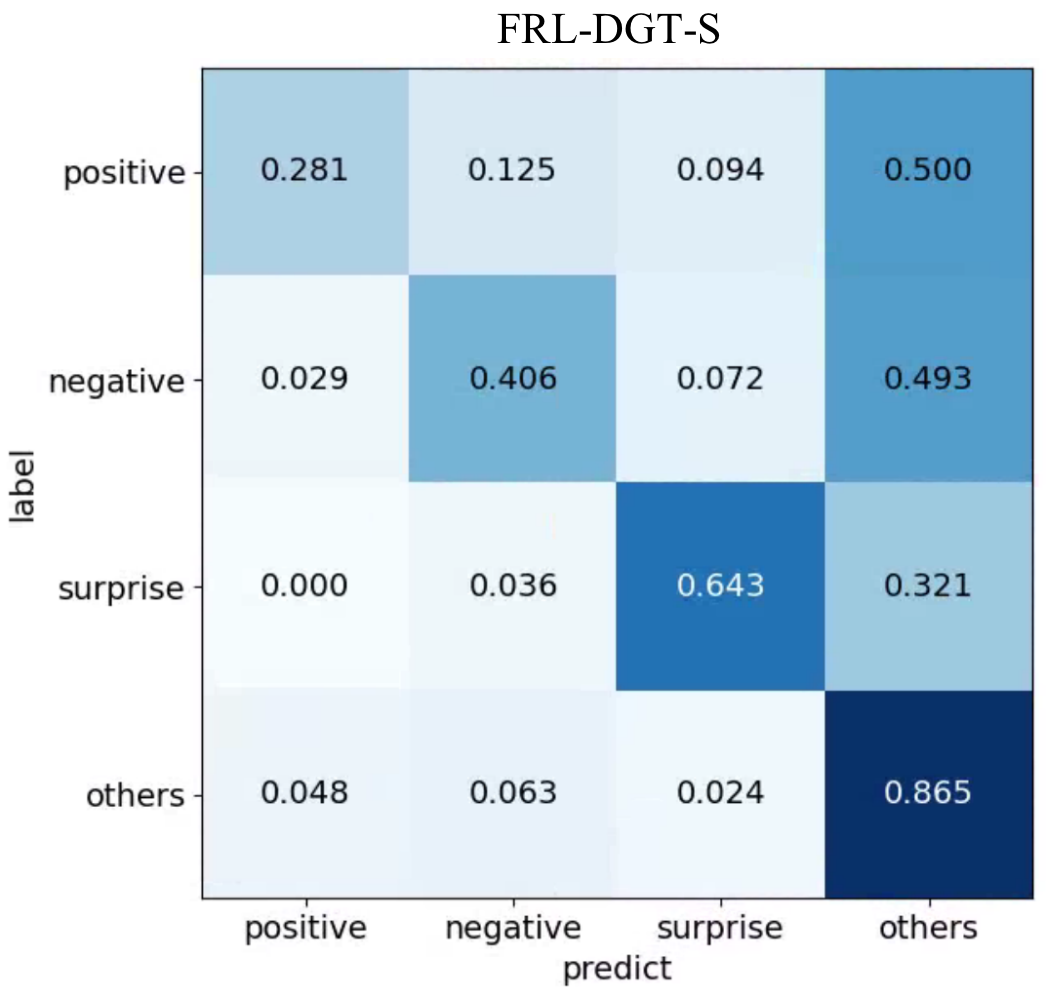}
     \hspace{-0.15cm}
     \includegraphics[height=0.22\textwidth]{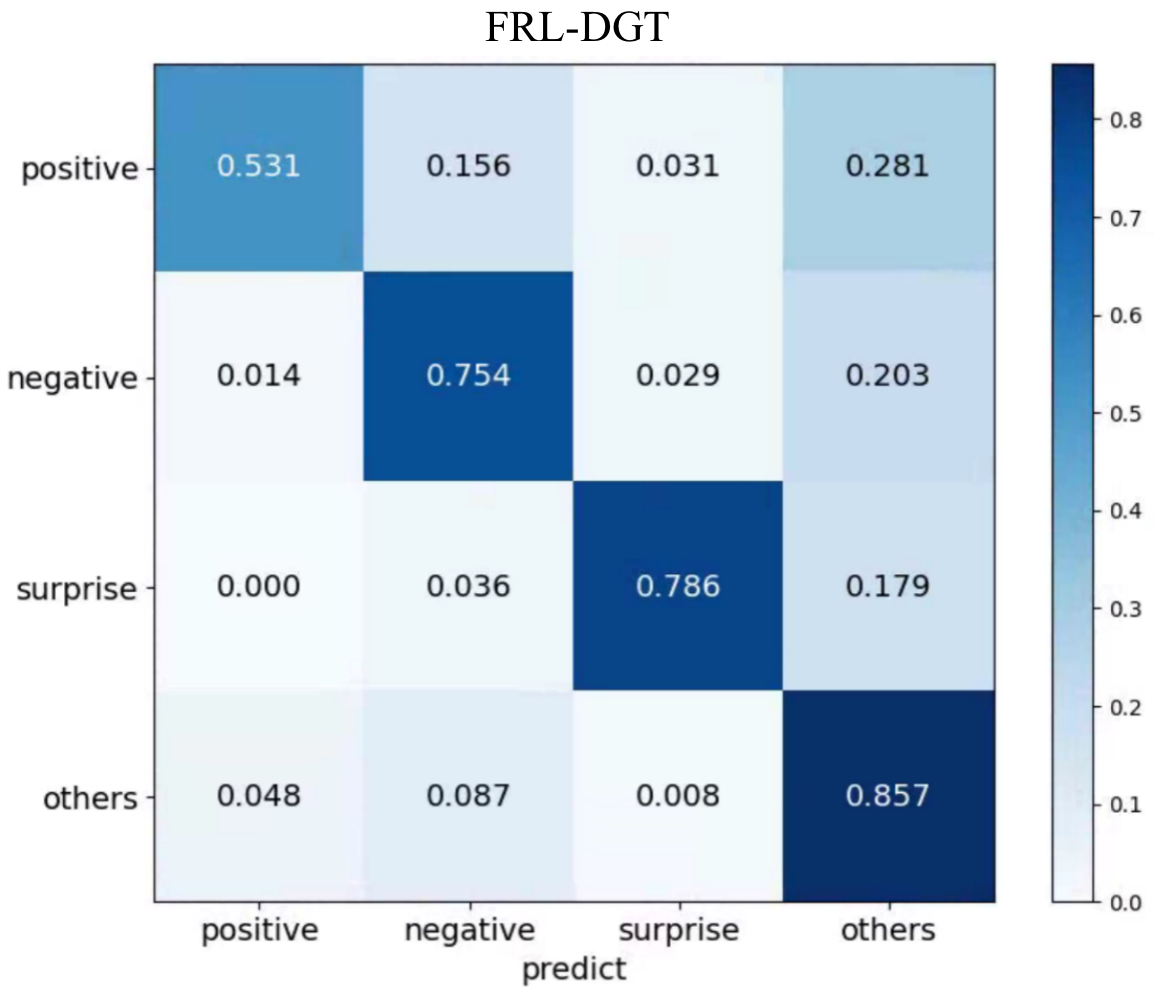}
     \hspace{-0.1cm} }
   \subfloat[5 classes]{
      \includegraphics[height=0.22\textwidth]{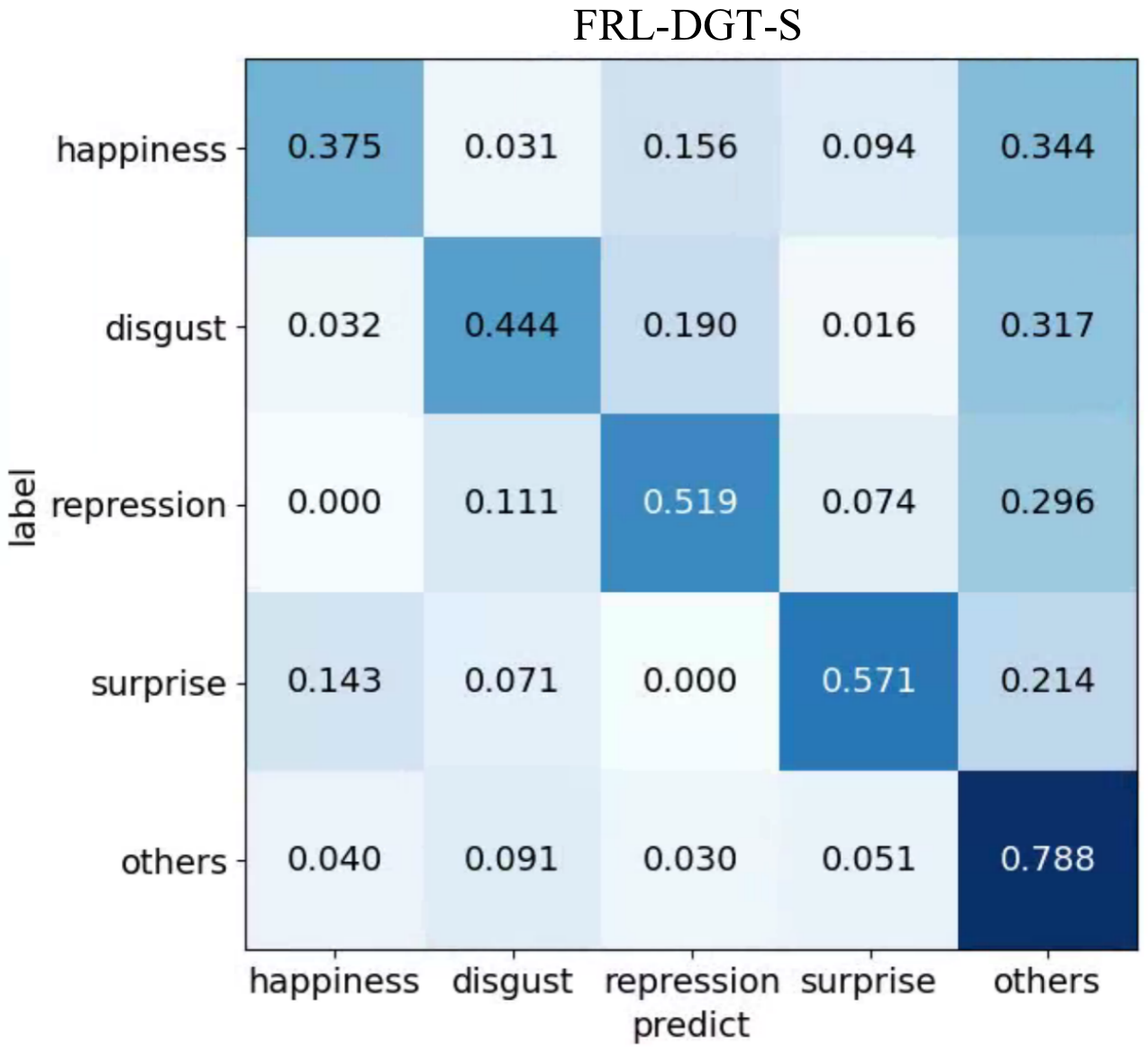} 
     \hspace{-0.15cm}
     \includegraphics[height=0.22\textwidth]{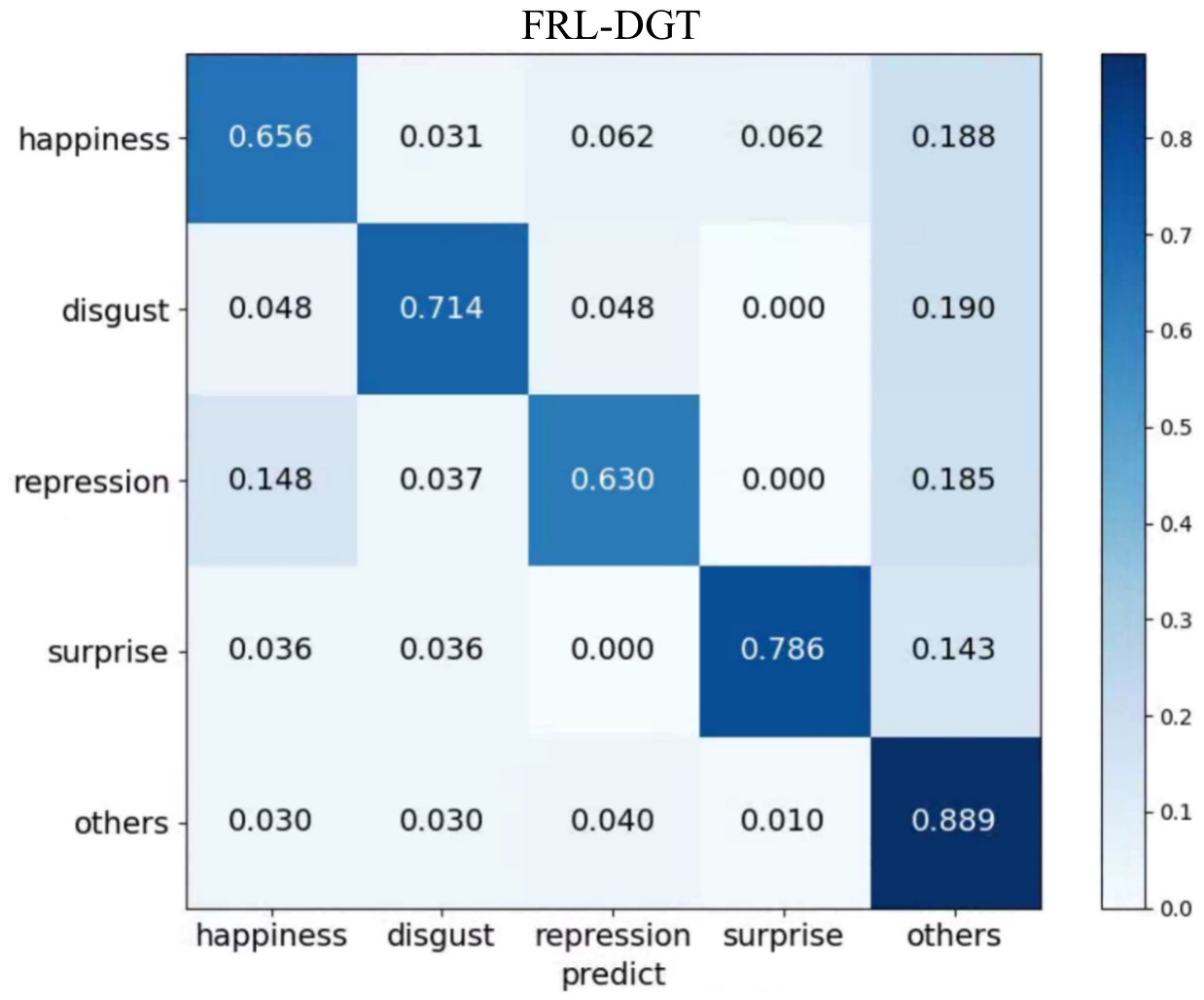} }
   \vspace{-0.35cm}
   \caption{The confusion matrices of our proposed FRL-DGT and FRL-DGT-S on CASME II dataset with 4 and 5 ME classes.}
   \label{fig:Confusion_matrices}
   \vspace{-0.30cm}
\end{figure*}


\textbf{AU Regions}. We explore the difference between using 9 different AUs and using 3×3 image patches divided evenly as input. From M5 and M9 shown in Table~\ref{tab:ablation experiments}, we can clearly see that using specific AUs allows the Transformer Fusion to perform more targeted learning, where the lower layers can focus on extracting features from individual AUs, while the higher layers can classify based on information of all AU regions, resulting in better performance. 

\subsection{Discussions}

\textbf{Extension of FRL-DGT on the Whole Sequence}.
For the classification of an expression sequence, we extend our FRL-DGT to FRL-DGT-S, in which the input data has an additional time dimension and the sequence is converted to $T$ frames by the Temporal-Interpolation-Model (TIM)~\cite{zhou2011towards}. The onset frame is concatenated with all other frames separately to obtain a sequence containing $T-1$ onset-other image pairs as input to DGM-S. Then the output displacement concatenated with its corresponding frame is input to sequenced Transformer Fusion (TransFu-S) for classification, and TransFu-S requires a further fusion step to model the temporal dependencies between $T$ frames in addition to the two-level fusion of TransFu. 
In our experiments, image sequences are interpolated to 10 frames using TIM, 
with the batch size 5 and same optimizer setting as FRL-DGT.

For the comparison of FRL-DGT-S and the SOTA methods which take image sequences as input, we conduct experiments on CASME II dataset. However, the classes used by ELRCN and STRCN-A are different, ELRCN uses 5 classes ({\em i.e.}, Happiness, Disgust, Repression, Surprise and Others) while STRCN-A uses 4 classes ({\em i.e.}, Positive, Negative, Surprise and Others), so we conduct experiments separately and compare with the corresponding methods. 

The comparison results are listed in Table~\ref{tab:comparative experiments for FRL-DGT-S}, where we cite the results from papers of ELRCN and STRCN-A, and 
our FRL-DGT-S outperforms them under the corresponding settings. We can also find from Table~\ref{tab:comparative experiments for FRL-DGT-S} that FRL-DGT has higher precision than FRL-DGT-S for both 4 classes and 5 classes, which is previously proved by related works and illustrated with confusion matrices in Figure~\ref{fig:Confusion_matrices}. 
\textcolor{black}{To investigate the reason why FRL-DGT-S is not as effective as FRL-DGT, we compare the interpolated frames after TIM and the original apex frame, which indicates that TIM may miss apex information, resulting in less good performance.}

\begin{table}[ht!]
  \centering
  \setlength{\abovecaptionskip}{10pt}
  \setlength{\tabcolsep}{3mm}{
  \resizebox{0.9\linewidth}{!}{\begin{tabular}{@{}c|l|c|ccc}
    \toprule
     \multirow{2}{0.02\textwidth}{} & \multirow{2}{0.08\textwidth}{Method} & \multirow{2}{0.08\textwidth}{\#(Classes)} &
    \multicolumn{3}{c}{CASME II} \\
    \cline{4-6}
    & & & UF1 &UAR & Acc \\
    \midrule
    \multirow{3}{0.01\textwidth}{\rotatebox{90}{Pair}} 
    & FRL-DGT & 4 & \textcolor{red}{0.750} & 0.732 & \textcolor{red}{0.780} \\
    & FRL-DGT & 5 & 0.748 &  \textcolor{red}{0.735} & 0.757 \\
    \midrule
   \multirow{4}{0.01\textwidth}{\rotatebox{90}{Sequence}} & STRCN-A~\cite{xia2019spatiotemporal} & 4 & 0.542 & - & 0.560 \\
    & FRL-DGT-S & 4	& \textcolor{red}{0.562}	& \textcolor{red}{0.549} & \textcolor{red}{0.643} \\
    \cline{2-6}
    & ELRCN~\cite{khor2018enriched} & 5 & 0.500 & 0.440 & 0.524 \\
    & FRL-DGT-S & 5 & \textcolor{red}{0.543} & \textcolor{red}{0.539} & \textcolor{red}{0.594} \\
    \bottomrule
  \end{tabular}}}
  \vspace{-0.25cm}
  \caption{Results of FRL-DGT on more ME classes and FRL-DGT-S taking image sequences as input.}
  \label{tab:comparative experiments for FRL-DGT-S}
  \vspace{-0.15cm}
\end{table}

\textbf{Sensitivity To Onset and Apex}.
Perturbations on onset/apex frames of CASME II include 10/20/30$\%$ deviation between onset and apex, e.g., onset+10$\%$ is about 3 frames after onset. 
\textcolor{black}{The averaged UF1/UAR from onset deviations are 0.848/0.838, 0.817/0.808, 0.786/0.769, while the results are 0.866/0.858, 0.830/0.834, 0.814/0.795 for apex.}

\textbf{Automatic Detection on Apex}.
\textcolor{black}{Without using the labeled frames, we automatically detect apex frames with the algorithm of~\cite{van2019capsulenet}, causing the averaged UF1/UAR decrease from 0.812/0.811 to 0.658/0.648 for all the three datasets.}

\textbf{Running Time}.
All experiments are performed on 3080 Ti GPU with 12GB memory, and i9 CPU with 2.8GHz. The average time of reading one picture from dataset, detecting the face, and outputting classification result is 0.416s, while the average time is 0.384s only for ME recognition.

\begin{figure}[t]
  \vspace{-0.2cm}
  \centering
    \includegraphics[width=0.155\linewidth]{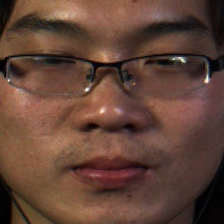}
    \includegraphics[width=0.155\linewidth]{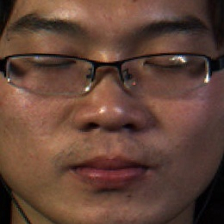}
    \includegraphics[width=0.16\linewidth]{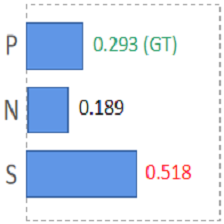} 
    \includegraphics[width=0.155\linewidth]{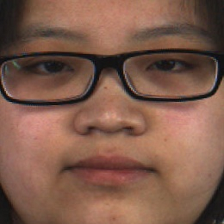}
   \includegraphics[width=0.155\linewidth]{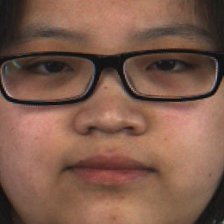}
   \includegraphics[width=0.16\linewidth]{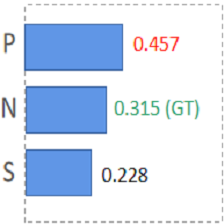}\\
   \vspace{-0.25cm}
   \caption{Failure cases of our FRL-DGT. GT stands for Ground Truth. (Images from CASME II \copyright{Xiaolan Fu})}
   \label{fig:failure_case}
   \vspace{-0.35cm}
\end{figure}

\textbf{Failure Cases}.
There are some failure cases that are difficult to be accurately classified by the existing methods and our FRL-DGT. As shown in Figure~\ref{fig:failure_case}, the glasses with reflection can confuse the extraction and classification of displacement features, leading to incorrect predictions.

\section{Conclusion}

For micro-expression recognition, we propose an end-to-end FRL-DGT, which takes onset-apex image pairs as input. The convolutional module DGM with self-supervised learning is used instead of traditional dynamic feature extractions, and the classification loss can back-propagate the gradient to DGM, modifying the information contained in the generated displacement. Our designed classification module Transformer Fusion consists of Transformer’s basic layers and the novel fusion layer, utilizing cropped AU regions and the full-face region as input for multi-level learning and fusion, and using the linear fusion before attention mechanism with efficient and accurate integration of embedding vectors. The LOSO evaluation result of our FRL-DGT has higher precision than the SOTA methods on UF1 and UAR tested with the same datasets, and the ablation experiments demonstrate the effectiveness of each proposed module.

\newpage
{\small
\bibliographystyle{ieee_fullname}
\bibliography{ref}
}

\end{document}